%% file: main.tex
\documentclass[10pt,twocolumn,letterpaper]{article}

\usepackage{cvpr}              
\input{preamble}
\definecolor{cvprblue}{rgb}{0.21,0.49,0.74}
\usepackage[pagebackref,breaklinks,colorlinks,allcolors=cvprblue]{hyperref}
\usepackage{multirow}

\def\paperID{} 
\def\confName{CVPR}
\def\confYear{2026}

\title{ForeHOI: Feed-forward 3D Object Reconstruction from Daily Hand-Object Interaction Videos}

\author{%
  \begin{minipage}{0.90\textwidth}
    \centering
    Yuantao Chen\textsuperscript{1$\ast$} \quad 
    Jiahao Chang\textsuperscript{2,1$\ast$} \quad
    Chongjie Ye\textsuperscript{2,1} \quad
    Chaoran Zhang\textsuperscript{4} \quad 
    Zhaojie Fang\textsuperscript{1} \\
    Chenghong Li\textsuperscript{2,1$\dagger$} \quad
    Xiaoguang Han\textsuperscript{1,2,3$\dagger$}\\[10pt]
    \textsuperscript{1}School of Science and Engineering, The Chinese University of Hong Kong, Shenzhen \quad
    \textsuperscript{2}FNii-Shenzhen \quad
    \textsuperscript{3}Guangdong Provincial Key Laboratory of Future Networks of Intelligence \quad
    \textsuperscript{4}SDS, CUHKSZ \\
  \end{minipage}
}

\begin{document}

\twocolumn[{%
\renewcommand\twocolumn[1][]{#1}%
\maketitle
\vspace{-10mm}
\begin{center}
    \centering
    \captionsetup{type=figure}
    \includegraphics[width=1.\textwidth]{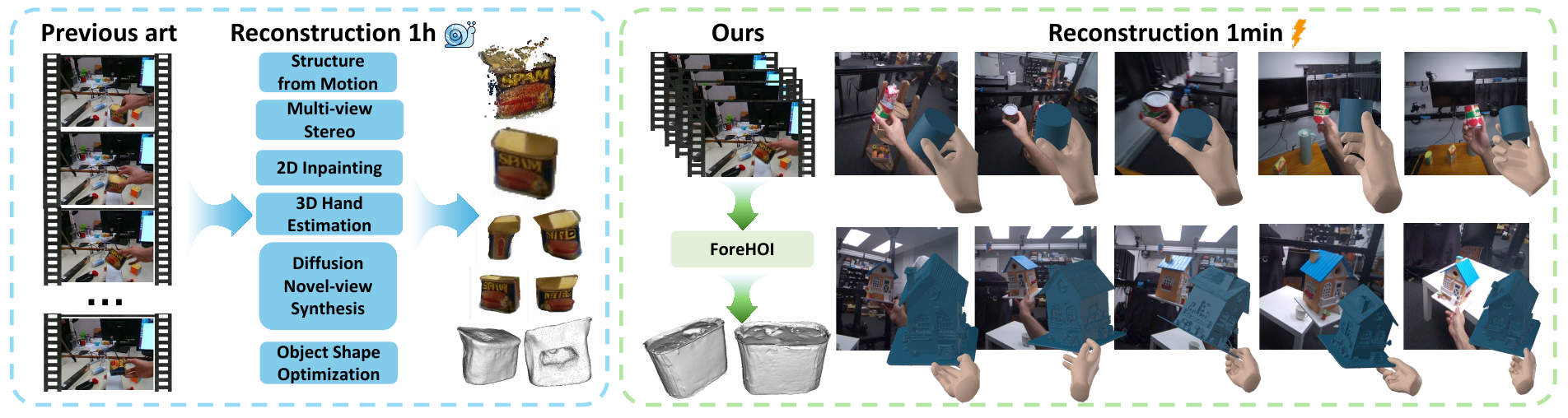}
    
    \captionsetup{skip=5pt} 
    \captionof{figure}{
    We propose \textit{ForeHOI}, the first feed-forward method that can directly reconstruct 3D object geometry from monocular hand-object interaction videos. Compared with previous methods that rely on complex pre-processing, our end-to-end pipeline achieves superior reconstruction performance under severe hand-object occlusion scenarios within one minute of inference time.
    \label{fig:teaser}}
\end{center}%
}]


\input{sec/0_abstract}    
\input{sec/1_intro}

\input{sec/2_related}

\input{sec/3_methods}
\input{sec/4_exp}
\input{sec/5_conclusion}

\section{Acknowledgments}
The work was supported in part by Guangdong S\&T Programme with Grant No. 2024B0101030002, the Basic Research Project No. HZQB-KCZYZ-2021067 of Hetao Shenzhen-HK S\&T Cooperation Zone, by Guangdong Provincial Outstanding Youth Fund with No. 2023B1515020055, the Shenzhen Outstanding Talents Training Fund 202002, the NSFC with Grant No. 62293482, the Guangdong Research Projects No. 2017ZT07X152 and No. 2019CX01X104, the Guangdong Provincial Key Laboratory of Future Networks of Intelligence (Grant No. 2022B1212010001), and the Shenzhen Key Laboratory of Big Data and Artificial Intelligence (Grant No. SYSPG20241211173853027), the Guangdong Province Radio Science Data Center.
{
    \small
    \bibliographystyle{ieeenat_fullname}
    \bibliography{main}
}

\end{document}

%% file: sec/0_abstract.tex

\begin{abstract}
The ubiquity of monocular videos capturing daily hand-object interactions presents a valuable resource for embodied intelligence. While 3D hand reconstruction from in-the-wild videos has seen significant progress, reconstructing the involved objects remains challenging due to severe occlusions and the complex, coupled motion of the camera, hands, and object. In this paper, we introduce ForeHOI, a novel feed-forward model that directly reconstructs 3D object geometry from monocular hand-object interaction videos within one minute of inference time, eliminating the need for any pre-processing steps. Our key insight is that, the joint prediction of 2D mask inpainting and 3D shape completion in a feed-forward framework can effectively address the problem of severe occlusion in monocular hand-held object videos, thereby achieving results that outperform the performance of  optimization-based methods. The information exchanges between the 2D and 3D shape completion boosts the overall reconstruction quality, enabling the framework to effectively handle severe hand-object occlusion. 
Furthermore, to support the training of our model, we contribute the first large-scale, high-fidelity synthetic dataset of hand-object interactions with comprehensive annotations. Extensive experiments demonstrate that ForeHOI achieves state-of-the-art performance in object reconstruction, significantly outperforming previous methods with around a 100x speedup. Code and data are available at: https://github.com/Tao-11-chen/ForeHOI.
\end{abstract}
\vspace{-20pt}

%% file: sec/1_intro.tex
\section{Introduction}
\label{sec:intro}

Every day, we unconsciously use our hands to interact with countless objects. With 3D hand-object interaction (HOI) data becoming increasingly important in embodied intelligence~\cite{zhang2024artigrasp, christen2022dgrasp, fu2025gigahands, chen2025web2grasp}, the comprehensive digitalization of 3D objects and hands  is essential. 3D hand reconstruction for daily monocular videos has been widely studied in the past few years~\cite{bib:handos, potamias2024wilor, ye2025predicting}. However, research on object target reconstruction from such video data has been neglected due to two key challenges: 1): Such daily-life video usually do not provide full observation of the object due to hand-induced occlusions and object self-occlusions; 2): The camera, object and hands are all dynamic in one monocular video, making it hard to estimate the relative motion between the object and camera. In this context, we define our task as the reconstruction of 3D object from daily hand-object interaction videos.

Since there exist unobserved areas for objects, this problem naturally needs strong object priors. As foundation models~\cite{wang2025vggt, liu2023zero1to3, xiang2024structured} become ever more powerful, recent contributions have tried to tackle this problem using priors from foundation models. EasyHOI~\cite{liu2025easyhoi} simplifies the problem to a single image input task and utilizes pretrained 2D inpainting models together with 3D generation models for complete object reconstruction. However, they are unable to utilize potential video sequence information, which is important for accurate 3D reconstruction in visible regions. MagicHOI~\cite{wang2024Magichoi} takes short video clips as input and integrates a large pretrained novel view synthesis model~\cite{liu2023zero1to3} into the object radiance field optimization pipeline for 3D object completion. However, its complex pipeline, including 2D inpainting and novel-view based 3D reconstruction, introduces view-inconsistent errors in each stage, leading to cumulative inaccuracies. Also, optimizing radiance fields usually takes hours, making it too slow for many applications. An alternative way for reconstructing hand-held objects involves first leveraging VGGT~\cite{wang2025vggt} for initial reconstruction, followed by a 3D completion method to achieve full object reconstruction. However, under severe hand-object occlusion scenarios, VGGT fails to produce satisfactory initial results, which consequently leads to poor performance in the final completion results.  Given the challenge of feed-forward reconstruction of severely occluded hand-held objects in monocular video, it is crucial to incorporate both 2D and 3D shape completion of the occluded regions.


In this paper, we take a further step toward removing pre-processing steps by introducing ForeHOI, a feed-forward 3D reconstruction model that takes a video clip as input and predicts object geometry within one minute, making the reconstruction process robust and easy to use for daily HOI videos. Our key insight is that, the joint prediction of 2D mask inpainting and 3D shape completion in a feed-forward framework can effectively address the problem of severe occlusion in monocular hand-held object videos. We demonstrate that with sufficient data, hand-held object reconstruction can be directly learned by a single neural network,  achieving results comparable with optimization-based methods but at a significantly higher inference speed.
 In detail, inspired by ReconViaGen~\cite{chang2025reconviagenaccuratemultiview3d}, we combine the reconstruction prior with a hand prior~\cite{potamias2024wilor} that provides the model with a deterministic understanding of the hand's shape and position as an additional condition. We then introduce a bidirectional cross-attention mechanism to the diffusion model to facilitate information exchange between the 3D features and the input image features. Through this bidirectional process,  the inpainted 2D object mask and the complete 3D object geometry are jointly estimated. The mutual refinement between them boosts the overall quality, enabling the framework to effectively handle severe hand-object occlusion.
 Furthermore, as no large-scale HOI dataset is available to train our model, we build a synthetic dataset based on GraspXL~\cite{zhang2024graspxl}: a robust RL-based grasp synthesis method. We first add photorealistic textures~\cite{HTML_eccv2020} to the MANO~\cite{10.1145/3130800.3130883} hands and then render multi-view videos of the synthesized grasping process using the Objaverse~\cite{objaverse} dataset. This pipeline ultimately yields a large-scale synthetic dataset of 400K samples with comprehensive annotations, including hand masks, object masks, hand poses, object poses, and depth maps.

Extensive experiments conducted on the widely-used  HO3D~\cite{hampali2020honnotate} and recent proposed challenging HOT3D~\cite{banerjee2024hot3d} datasets demonstrate that ForeHOI achieves state-of-the-art performance in  object geometry reconstruction, with significantly faster speed than optimization-based methods. Our contributions are summarized as follows: 
\begin{itemize}
[itemsep=2pt,topsep=2pt,parsep=0pt, leftmargin=1.5em]
    \item We propose ForeHOI, the first feed-forward method that effectively predicts object geometry from daily monocular HOI videos within one minute of inference time.
    \item  We jointly generate 2D object mask inpainting and 3D object completion within our framework to address the reconstruction challenge provided by severe hand-object occlusion in daily monocular HOI videos.
    \item We contribute the first large-scale synthetic hand-object video dataset with high-fidelity rendering images.
\end{itemize}

%% file: sec/2_related.tex
\section{Related Work}
\label{sec:related}

\textbf{Object Reconstruction from generative priors.} Preceding studies~\cite{DBLP:journals/corr/abs-2310-15110, watson2022novelviewsynthesisdiffusion, ICLR2024_5e8309c9, Hollein_2024_CVPR, Wu_2024_CVPR, liu2023syncdreamer} represented by Zero-1-to-3~\cite{liu2023zero1to3} focus mainly on leveraging 2D generative priors~\cite{rombach2021highresolution, NEURIPS2022_ec795aea} into 3D object reconstruction pipeline. These methods suffer from multi-view inconsistency problems and usually have lower quality compared with native 3D generation models~\cite{hunyuan3d2025hunyuan3d, xiang2024structured, kang2024robin3d, TripoSR2024, li2025triposg} on single-image input tasks.  The very recent attempts CUPID~\cite{huang2025cupidposegroundedgenerative3d} and ReconViaGen~\cite{chang2025reconviagenaccuratemultiview3d} leverage native 3D generation models for 3d reconstruction tasks by introducing pose-grounded generation and reconstruction priors, achieving superior performance on the single-image reconstruction task. Our work focuses on video input reconstruction under HOI scenarios, with a different approach, but leads to similar conclusions.

\noindent\textbf{3D Hand Reconstruction.} Over the past few decades, significant progress has been made in 3D hand reconstruction technology.
With the introduction of the parametric hand model MANO~\cite{10.1145/3130800.3130883}, recent studies~\cite{Baek_2019_CVPR, Hampali_2022_CVPR_Kypt_Trans, Kulon_2020_CVPR, Lin_2021_CVPR, Pavlakos_2024_CVPR, zhang2020phosa, Liu_2021_CVPR} have integrated it as a differentiable module into neural networks, enabling end-to-end reconstruction of dense 3D hand meshes directly from images or videos. State-of-the-art methods~\cite{bib:handos, potamias2024wilor, ye2025predicting} driven by large-scale data not only achieve robust 3D hand reconstruction performance but also demonstrate strong generalization capabilities to in-the-wild images and videos. We use end-to-end hand reconstruction as guidance to improve object reconstruction.

\noindent\textbf{Template-based Hand-held Object Reconstruction.} Hand-held object reconstruction is more challenging than hands due to the object shape's natural diversity and hand-induced occlusion. Some existing methods simplify this problem by taking object templates as input~\cite{Cao_2021_ICCV, Corona_2020_CVPR, fan2024benchmarks, liu2021semi, Tekin_2019_CVPR, Yang_2021_ICCV, Yang_2022_CVPR} and focusing mainly on estimating 6-DoF object pose. A recent contribution, Dynhor~\cite{jiang2025hand} tackles the problem by generating the object template using a text-prompted 3D generation model with textual description coming from ChatGPT. Such a process creates large uncertainty due to the limitation of textual prompts. In contrast, our work predicts the object pose and shape simultaneously.

\noindent\textbf{Template-free Hand-held Object Reconstruction.} Recent advancements have introduced more generalizable methods~\cite{Ye_2024_CVPR, Ye_2023_ICCV, bundlesdfwen2023, Wang_2025_ICCV, fan2024hold, huang2022hhor, Prakash2024HOI, Qu_2023_ICCV, Hasson_2019_CVPR, wu2024reconstructing, jiang2024hand} with the development of general object reconstruction techniques like implicit fields~\cite{mildenhall2020nerf, wang2021neus} and Diffusion models~\cite{poole2022dreamfusion}. However, these methods often impose strong constraints, including depth input~\cite{bundlesdfwen2023}, accurate object pose input~\cite{Wang_2025_ICCV, fan2024hold}, rigid hand-object interactions~\cite{huang2022hhor, Prakash2024HOI}, multi-view observations~\cite{Qu_2023_ICCV}, category-level optimization ~\cite{Hasson_2019_CVPR, wu2024reconstructing}, and complete object observations~\cite{fan2024hold}. Thus, they are difficult to work on daily hand-object interaction videos. On the contrary, our method takes only RGB video clips as input and ensures 3D reconstruction quality with a native 3D Diffusion model. Another branch of research~\cite{Chen_2023_CVPR, chen2025hort, liu2025easyhoi, aytekin2025followholdhandobjectinteraction, NEURIPS2023_b2876deb, chen2022alignsdf, Ye_2022_CVPR} simplifies the problem to a single-image input task. Although they minimize input requirements, they are unable to consider potential video data, leading to an inability to exploit the temporal and multi-view information inherent in video sequences.

\noindent\textbf{Hand-object interaction datasets.} The lack of hand-object interaction data is a major bottleneck for the generalization capability of generalizable object reconstruction methods. Although several datasets have been proposed~\cite{hasson19_obman, hampali2022keypointtransformer, hampali2020honnotate, li2025vitra, fu2025gigahands, Liu_2022_CVPR, Jian_2023_ICCV} in previous research, the largest real-world dataset~\cite{Liu_2022_CVPR} currently available contains only 417 objects, while the largest synthetic dataset~\cite{hasson19_obman} comprises 2772 objects. Such a limited scale is far from sufficient for modeling complex object shapes. A recent contribution, GraspXL~\cite{zhang2024graspxl} introduced a robust RL-based method for grasping synthesis and released their synthesizing results on Objaverse~\cite{objaverse}. We thus built the first large-scale hand-object interaction dataset based on it with more than 400k video sequences.

%% file: sec/3_methods.tex
\section{Methods}
\label{sec:methods}

\begin{figure*}[htb]
  \centering
   \includegraphics[width=1.0\linewidth]{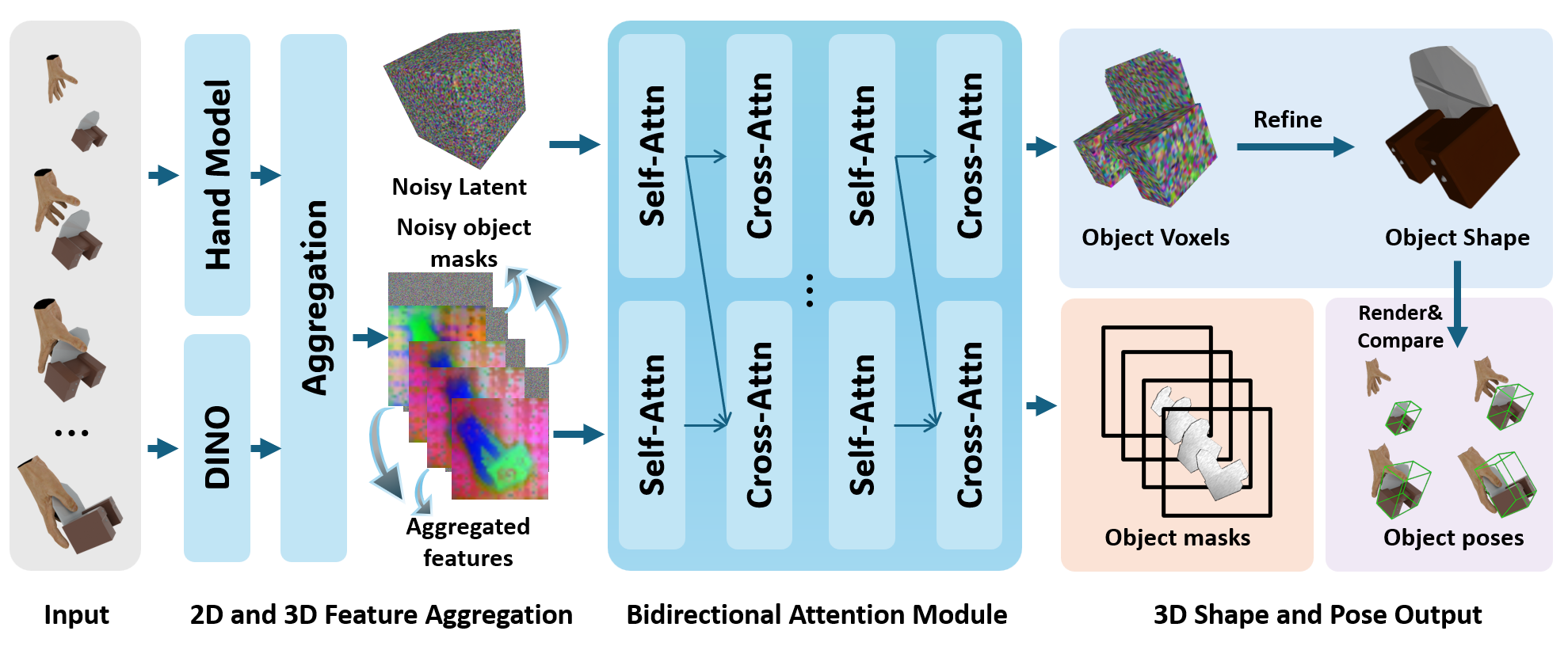}   
   \caption{\textbf{Pipeline overview of the proposed ForeHOI.} Given a monocular video of hand-object interaction, we adopt a diffusion-based framework that jointly performs 2D object mask inpainting and 3D object completion to address the reconstruction challenge posed by severe hand-object occlusion. Moreover, the accurate object shape reconstruction achieved by our method leads to precise 3D object pose estimation through post-processing.}
   \label{fig:main}
   \vspace{-2mm}
\end{figure*}



Figure~\ref{fig:main} gives an overview of our method, ForeHOI, which reconstructs the object geometry from an RGB video sequence with limited viewpoints in a feed-forward manner. We begin by describing our image and hand feature encoding schemes in Sec.~\ref{method31}. Then Sec.~\ref{method32} details a bidirectional cross-attention mechanism inside a diffusion transformer (DiT) that establishes correspondences between object voxels and multi-view masks. 
The training and inference process are detailed in Sec.~\ref{method33}.
With the recovered mesh, per-frame object poses are obtained via rendering and comparison (Sec.~\ref{method34}). Finally, Sec.~\ref{method35} introduces our new large-scale dataset.

\subsection{Image and hand prior encoding}
\label{method31}
 We use DINOv2~\cite{oquab2023dinov2} as the image feature extractor to encode each frame of the input video. Apart from image features, we also encode hand features using a state-of-the-art hand pose estimation model~\cite{potamias2024wilor} for each image as input. Specifically, the image patch features of the hand ViT backbone $F_{hand} \in \mathbb{R}^{n\times1024}$ are aggregated with the DINOv2 image patch features in a patch-to-patch manner for each input image into $F_{input} \in \mathbb{R}^{n\times2048}$, finally we project it back to $\mathbb{R}^{n\times1024}$ with a two-layer MLP. 
 
 By using the feature aggregator module to aggregate the 2D and 3D hand features, the model will have a deterministic understanding of the hand's shape and position. Note that since the features are pixel-aligned with each image, they contain implicit information in a local camera space, eliminating the multi-frame scale inconsistency problem. 
 By using such an encoding strategy, we minimize the input requirements: no separate hand or object mask is needed during inference.

\subsection{Bidirectional cross attention}
\label{method32}
Our ultimate goal is to recover the object’s complete shape. However, our early experiments show that directly predicting an accurate and complete geometry is highly challenging. We therefore decompose the problem into two complementary stages: 2D and 3D shape completion.
Because the voxel-generation stage lacks sufficient appearance cues, we train the backbone with two objectives: per-frame pixel-aligned multi-view complete masks and the object’s complete voxel representation. These two objectives are tightly coupled and mutually reinforcing. We elaborate on the benefits in both directions, beginning with object-voxel estimation: each input 2D image contains an incomplete observation of the object due to hand-induced occlusions. Under such circumstances, the 3D shape generation backbone must hallucinate the occluded regions without direct 2D evidence. Supervising the model with complete 2D object masks provides additional guidance on the full object contour in each view, thereby enhancing its ability to infer a coherent and complete 3D shape.

Given that our two tasks are highly interdependent and mutually reinforcing, effective information exchange between the two branches is essential. To this end, we adopt a dual-branch architecture equipped with bidirectional cross-attention. Starting from a noisy latent $\in \mathbb{R}^{64\times64\times64}$, the geometry branch progressively denoises it into a coarse object structure conditioned on the input features. Following ReconViaGen~\cite{chang2025reconviagenaccuratemultiview3d}, we apply cross-attention between multi-view aggregated image features and the 3D latent representation within each DiT block using a weighted fusion strategy.
Differently, we replace the original input features with features from the object-mask branch—layer by layer from shallow to deep—so that the mask branch directly guides and supports the shape reconstruction process.
Just as mask-branch features are crucial for the geometry branch, features from the geometry branch are equally important for completing object masks. Inspired by the seminal multimodal alignment method~\cite{lu2019vilbert},  we treat the 3D features as contextual signals and feed them back into the mask branch via additional cross-attention. The whole process can be formulated as:
\begin{equation}
    \begin{split}
       y_{j+1} = \sum_{k=1}^{N}Cross&Attn(Q(y_{j}), K(x_{i}^{n}), V(x_{i}^{n}))\cdot w_{n}, \\
       x_{i+1} = Cross&Attn(Q(x_{i}), K(y_{j}), V(y_{j})), \\
       j \in \{m\}_{m=1}^{M}, &i \in \{p\}_{p=1}^{P},k \in \{n\}_{n=1}^{N}.
    \end{split}
\end{equation}
where $M$ is the number of DiT blocks in the geometry branch, $P$ is the number of DiT blocks in the mask branch, $N$ is the number of frames, $y_{j}$ denotes the 3D feature output of the $j$-th geometry branch block, while $x_{i}$ denotes the batch of image features after the $i$-th block of the mask branch. $w_{n}$ is the fusion weight associated with the $n$-th frame.
Notably, because multi-view image features are fused through weighted cross-attention rather than concatenation, the model can accommodate relatively long input sequences with only a modest increase in memory consumption—unlike approaches that rely on concatenated multi-image features~\cite{wang2025vggt, hunyuan3d22025tencent}.

\subsection{Training and inference}
\label{method33}
We train the network end-to-end by minimizing the conditional flow matching (CFM) objective for both 3D latent and 2D masks simultaneously with the same random $t$ at each step, with the following loss:
\begin{equation}
\begin{split}
    \mathcal{L}_{CFM}(\theta) &= \mathbb{E}_{t,x_{0},\epsilon}||v_{\theta}(x,t) - (\epsilon - x_{0})||_{2}^{2}, \\
    \mathcal{L} &= \mathcal{L}_{CFM}^{2D} + \beta\mathcal{L}_{CFM}^{3D}.
\end{split}
\end{equation} 
where $t$ is the continuous time step uniformly sampled from $\mathcal{U}(0,1)$. $x_0$ represents the ground-truth data sample, and $\epsilon$ is the noise sampled from a standard Gaussian prior distribution $\mathcal{N}(0, I)$. $x$ denotes the interpolated intermediate state at time $t$. Accordingly, $v_{\theta}(x,t)$ is the velocity field predicted by the network, which is supervised by the ground-truth vector field $(\epsilon - x_0)$ pointing from the data to the noise. Furthermore, $\mathcal{L}_{CFM}^{2D}$ and $\mathcal{L}_{CFM}^{3D}$ denote the conditional flow matching losses for the 2D mask and 3D shape, $\beta$ is used to balance the loss of 2D mask and 3D shape.

To transform the voxel representation into a high-fidelity object surface, we train a masked multi-view structured-latent (SLat) flow following the strategy of ReconViaGen~\cite{chang2025reconviagenaccuratemultiview3d}. Unlike the original formulation, we replace the complete object images with occluded object images as inputs, and we directly adopt DINOv2 features instead of VGGT features, as the latter are unreliable when significant occlusions in object are present.
The loss function is the same as the second stage of TRELLIS~\cite{xiang2024structured}.

At inference time, the multi-view object mask and object shape are simultaneously denoised through the CFM's denoising process, given 3D noisy latent volume or 2D noisy latent image as $x_{0} \sim p_{0}$, the object's 3d shape and masks are denoised by the learned vector field $v(x,t) = \nabla_{t}x$ with Eular sampling methods: 
\begin{equation}
    x(1) = x(0) + \int_{0}^{1} v_{\theta}(x(t),t)dt.
\end{equation}
After object voxel generation, we elaborate its fined surface through masked multi-view structured-latent (SLat) flow finetuned on our data and decode it to 3D mesh following TRELLIS~\cite{xiang2024structured}. 

\subsection{Object pose estimation}
\label{method34}
Since a large number of objects in real-world scenarios exhibit a high degree of symmetry, a general approach that relies solely on geometry for object pose estimation is not robust: symmetric objects introduce rotational ambiguity. Therefore, we adopt a pose estimation method based on textures derived from generated 3D models. The texture generation stage of the model is also trained on our dataset; further details on texture quality can be found in the appendix.

We employ a render-and-compare strategy to align all input views to the object coordinate system. Concretely, we begin by rendering 30 reference images from camera viewpoints uniformly distributed on a sphere. These rendered images are concatenated with the input frames and passed through VGGT~\cite{wang2025vggt} to obtain coarse camera poses. Because the poses of the rendered views are known, we can derive the transformation that maps the predicted poses of VGGT~\cite{wang2025vggt} into the object space, thereby yielding an initial pose estimation for each input image.

We then refine these initial poses through iterative correspondence-based alignment. Specifically, we render images using the coarse poses and apply Mast3R~\cite{mast3r_arxiv24} to establish 2D correspondences between each input frame and its rendered counterpart. We use Mast3R~\cite{mast3r_arxiv24} because it is trained with abundant data augmentation and is robust to our generated textures, which is not perfect. The coarse camera poses are then refined using a PnP solver with RANSAC. After several rounds of this refinement procedure, the initial predictions are substantially improved, resulting in highly accurate camera pose estimates.

\subsection{Dataset building}
\label{method35}
To build the large-scale high-fidelity synthetic dataset, we first incorporate a parametric hand texture model~\cite{HTML_eccv2020} into the MANO~\cite{10.1145/3130800.3130883} hands, and randomize hand skin tone during rendering by adjusting its parameters. Then we carefully selected high-quality GraspXL synthesized grasping sequences on all the objects from Objaverse~\cite{objaverse} datasets by taking the object's texture quality and the object's degree of movement into consideration. Finally, we render the grasping processes as multi-view video sequences using Blender~\cite{blender}. During rendering, we automatically generate the camera moving trajectory for each grasping sequence that ensures the object and hand remain within the camera frustum at all times, while allowing for a limited range of object movement inside the rendered images during manipulation. Furthermore, Light intensity and direction are also randomized during rendering.

%% file: sec/4_exp.tex
\section{Experiments}
\label{sec:exp}

We conduct experiments to effectively evaluate our model against state‑of‑the‑art methods on the task of partially visible object reconstruction from daily hand-object-interaction videos. The objective is to recover both 3D object geometry accurately and the object pose from short monocular video sequences that provide only limited observation. 
Sec.~\ref{sec:implementation} outlines the implementation details and training configurations of our framework. Sec.~\ref{sec:dataset} introduces the datasets used for training and evaluation, including both existing benchmarks and our newly constructed dataset. Sec.~\ref{sec:metrics} summarizes the evaluation metrics employed to assess 3D reconstruction quality, 2D prediction accuracy, and camera pose estimation. Sec.~\ref{sec:quantitative} presents quantitative and qualitative comparisons with recent state-of-the-art methods across multiple settings. Finally, Sec.~\ref{sec:ablation} provides detailed ablation experiments to analyze the contribution of each component in our pipeline and to validate the design choices.

\begin{figure*}[htb]
  \centering
   \includegraphics[width=1.0\linewidth]{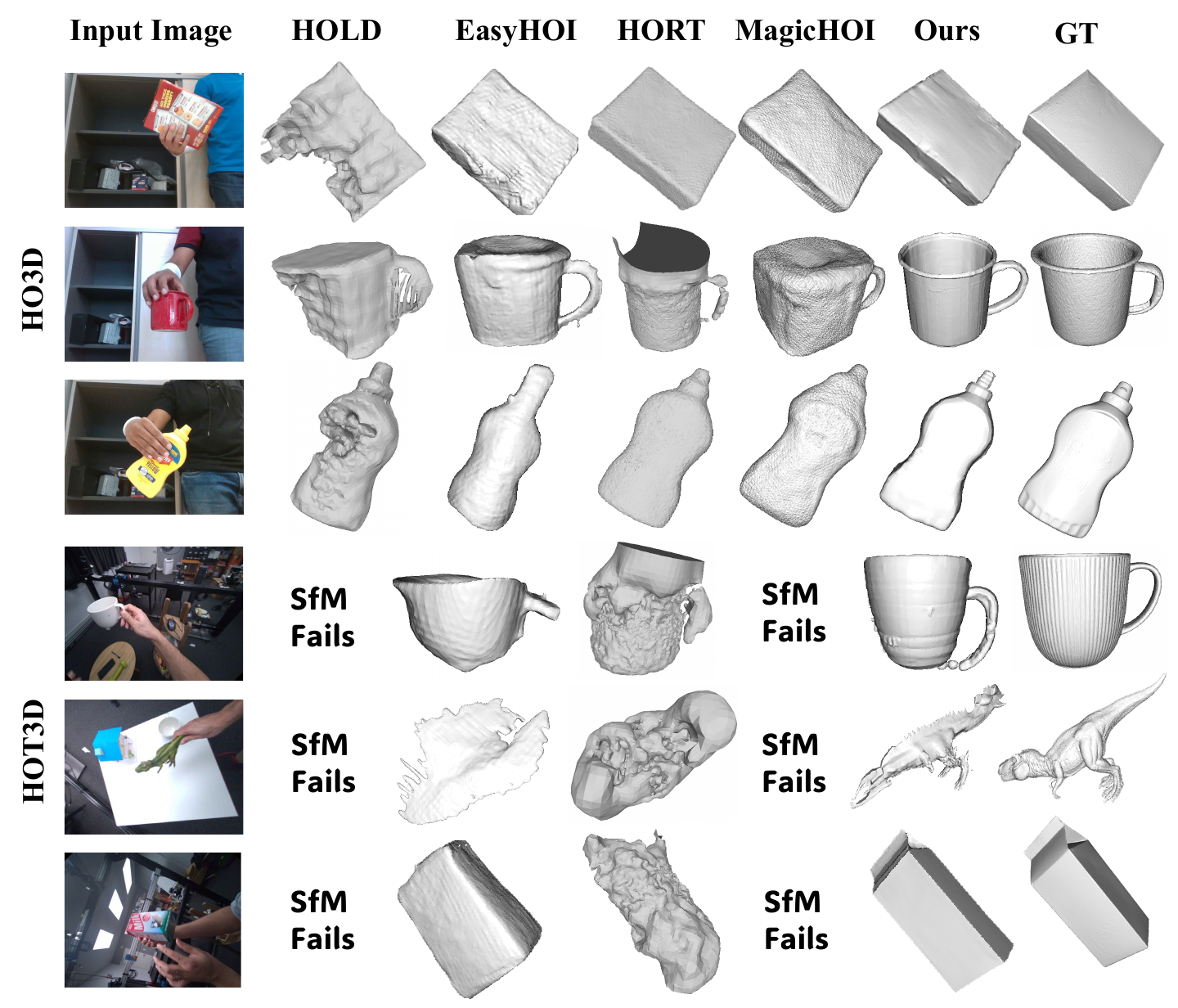}   
   \caption{\textbf{Qualitative visual comparison results on the HO3D ~\cite{hampali2020honnotate} and HOT3D~\cite{banerjee2024hot3d} datasets. SfM Fails represent the failure camera estimation from the structure-from-motion (SfM) method COLMAP since the sparse-view in HOT3D video clips. }}
   \label{fig:quali}
   \vspace{-2mm}
\end{figure*}


\subsection{Implementation Details}
\label{sec:implementation}
We adopt the same architecture as TRELLIS~\cite{xiang2024structured}.
We fine-tune the diffusion transformer in both stages of TRELLIS using LoRA. In our configuration, the LoRA rank is set to 64, the scaling parameter (alpha) is set to 128, and the dropout for all LoRA modules is set to 0. The LoRA adapters are inserted exclusively into the qkv projection layers and the output projection layers of each attention block.
The number of input views is randomly set as 2-6 while training, all the input images are resized to 518 $\times$ 518 images, resulting a 37 $\times$ 37 patches of image features, while the hand feature extractor takes 224 $\times$ 224 images as input, which is more compression, we then directly use a CNN with an MLP to fuse the two features as patch-level aggregated features. We randomly mix the images from different viewpoints at different timestamps as data augmentation.  During training, we employ the Adam optimizer with a learning rate of 5e-4, the training takes 4 days on 8 NVIDIA L20 40G GPUs with a total batch size of 32.

\subsection{Datasets}
\label{sec:dataset}
\noindent\textbf{Training data} We train our model only on our own large-scale synthetic dataset without any finetuning on real datasets, which highlights the strong data prior in our model and the importance of our dataset. Our data consists of over 400K video clips capturing hand-object interaction scenes in a multi-view manner. We randomly mix multi-view images with different timestamps on the same grasping sequence and randomly rotate or flip the input images as data enhancement while training.

\noindent\textbf{HO3D} We test our model on the HO3D~\cite{hampali2020honnotate}, a commonly used dataset for hand-object interaction reconstruction tasks. Specifically, we use the same sequences of HO3D-v3 as HOLD~\cite{fan2024hold}, resulting in a total of 14 sequences. Since we are focusing on reconstructing objects from data with limited observations, following previous art MagicHOI~\cite{wang2024Magichoi}, we divide long sequences into 30-frame clips. For single-view reconstruction methods, HoRT~\cite{chen2025hort} and EasyHOI, we test them on each image and average the final metrics for all views of a sequence.

\noindent\textbf{HOT3D} Since our motivation is to digitize 3D objects from daily hand-object interaction videos, among which, egocentric data is the most challenging. HOT3D~\cite{banerjee2024hot3d} is a recent dataset captured with VR headsets. Different from other common datasets~\cite{hasson19_obman, hampali2022keypointtransformer, hampali2020honnotate, li2025vitra, fu2025gigahands, Liu_2022_CVPR, Jian_2023_ICCV} that intentionally rotate the object in hand for reconstruction, it is the first dataset that truly captures humans' daily interaction with objects: we often interact with many objects at a short time and most of the interaction for one object happens only in 10-20 video frames, with highly dynamic and fast hand occlusions. To use the dataset for evaluation, we carefully preprocess the video, including cropping the video into short videos that describe the hand interaction with one object, undistorting the video frames, and sometimes cropping the video frames so that the object won't be too small in the video frame. Finally, we select 6 short clips at a length of 5-15 frames as evaluation data. 

\subsection{Evaluation metrics}
\label{sec:metrics}
Since we focus mainly on the object's reconstruction, following~\cite{chen2025hort, wang2024Magichoi, Jiang_2025_CVPR}, we use the Chamfer distance (CD) in centimeters to assess object reconstruction quality, however, it is highly sensitive to outliers, so we also compute the F-score in percentage at 5mm (F5) and 10mm (F10) for a comprehensive evaluation of the object's shape. Additionally, we evaluate pose accuracy using standard visual odometry metrics~\cite{6096039}, including Absolute Trajectory Error (ATE) and Relative Pose Error (RPE). In which ATE measures the absolute difference between the estimated and ground truth camera translation, which is the object's translation in camera space, in our case, while RPE represents the relative rotation ($RPE_{r}$) and translation ($RPE_{t}$) error. 

\subsection{Comparison with state of the art}
\label{sec:quantitative}
\textbf{Reconstruction results} We compare our object shape reconstruction quality with state-of-the-art (SOTA) hand-held object reconstruction methods in Tab.~\ref{tab:main table}, including HOLD~\cite{fan2024hold}, EasyHOI~\cite{liu2025easyhoi}, HORT~\cite{chen2025hort}, and MagicHOI~\cite{wang2024Magichoi}. Among them, HORT is trained on the HO3D dataset, so we only compare it on the HOT3D dataset. As shown in Fig.~\ref{fig:quali}, due to a lack of data, HORT only works well on similar objects from their limited training data and is hard to generalize to unseen objects from HOT3D, it even completely fails for relatively complex geometry or unusual camera viewpoints. HOLD is the only non-prior method, without an object shape prior, it's hard for it to reconstruct the complete object shape. Although EasyHOI utilizes rich priors from 2D inpainting models and 3D generation models, due to the accumulative error brought by its complex pipeline, the final 3D shape actually deviates from the original image a lot, and it also fails in line 5 of Fig.~\ref{fig:quali} for complex geometry. By utilizing 2D novel-view-synthesis priors, MagicHOI can not only complete the occlusion area, but also align well with the image in observed areas, however, due to the natural inconsistency of 2D multi-view generation model, the generated back of the object may have strange shapes, see line 2 in Fig.~\ref{fig:quali}, also, such an inconsistency makes the optimized surface quite rough and bumpy. Noting that both HOLD and MagicHOI need the object pose and sparse points from Structure-from-Motion (SfM) as input, thus, the whole pipeline fails whenever SfM fails, which is quite common in daily HOI scenes.

\begin{table}[htbp]
  \centering
    \begin{tabular}{c|ccc}
    \toprule
    \multicolumn{1}{l|}{Methods} & RPE(cm) & RPE(°) & ATE(m) \\
    \midrule
    HOLD  & 5.87  & 6.24  & 0.27 \\
    Dynhor & 4.25  & 5.25  & 0.24 \\
    Ours  & \textbf{1.42}  & \textbf{2.64}  & \textbf{0.13} \\
    \bottomrule
    \end{tabular}%
    \caption{Object pose estimation results compared with the HOLD and Dynhor method.}
  \label{tab:pose}%
  \vspace{-5mm}
\end{table}%

\textbf{Object pose results} We compare our object pose with Dynhor~\cite{Jiang_2025_CVPR} and HOLD~\cite{fan2024hold} in Tab.~\ref{tab:pose} on the HO3D dataset. Since HOLD actually takes Structure-from-Motion (SfM) poses as an initial pose and refines it, the quality of the initial pose matters for the final object pose. In our setting, the input is a video clip with a somewhat sparse viewpoint for the object, such a pattern is quite challenging for SfM methods, thus resulting in relatively poor pose estimation results. On the other hand, our method utilizes the reconstructed model for pose estimation, which is a template-based object estimation task, with the help of Mast3R's robust matching, the result turns out to be quite good. Dynhor claims to utilized a texture-to-3d generated template as input, we follow their paper to use Genie for shape generation, due to lack of text prompt of GPT's image capturing process, the resulting textured mesh has a high degree of randomness, thus influences the pose results, though they claimed that the following refining integrated in object reconstruction eliminates the errors in initial pose guessing process, this part is not open-sourced yet, so we use their pose at first step for comparison, which is not as good as the claim in their paper. 

\begin{table*}[htbp]
  \centering
    \begin{tabular}{cccccccc}
    \toprule
    \multirow{2}[4]{*}{Methods} & \multicolumn{3}{c}{HO3D} & \multicolumn{3}{c}{HOT3D} & \multirow{2}[4]{*}{Avg Time(min)} \\
\cmidrule{2-7}          & CD(cm) $\downarrow$ & \textcolor[rgb]{ .02,  .388,  .757}{F@5(\%) $\uparrow$} & \textcolor[rgb]{ .02,  .388,  .757}{F@10(\%) $\uparrow$} & CD(cm) $\downarrow$ & \textcolor[rgb]{ .02,  .388,  .757}{F@5(\%) $\uparrow$} & \textcolor[rgb]{ .02,  .388,  .757}{F@10(\%) $\uparrow$} &  \\
    \midrule
    EasyHOI & 1.83  & 46.35 & 69.24 & 1.21  & 18.46 & 34.25 & 175 \\
    HOLD  & 1.36  & 66.42 & 82.43 & N/A   & N/A   & N/A   & 330 \\
    HORT  & N/A   & N/A   & N/A   & 2.32  & 16.52 & 20.43 & \textbf{0.8} \\
    MagicHOI & 0.86  & 64.53 & 91.87 & N/A   & N/A   & N/A   & 58 \\
    Ours  & \textbf{0.79} & \textbf{68.95} & \textbf{93.72} & \textbf{1.03} & \textbf{60.5} & \textbf{89.1} & 1.1 \\
    \bottomrule
    \end{tabular}%
  \caption{Quantitative comparison with baseline methods on the real-world HO3D and HOI dataset. Our method outperform all the baseline method on geometry accuracy and inference time.}
  \label{tab:main table}%
  \vspace{-2mm}
\end{table*}%

\begin{figure}[htb]
  \centering
   \includegraphics[width=1.0\linewidth]{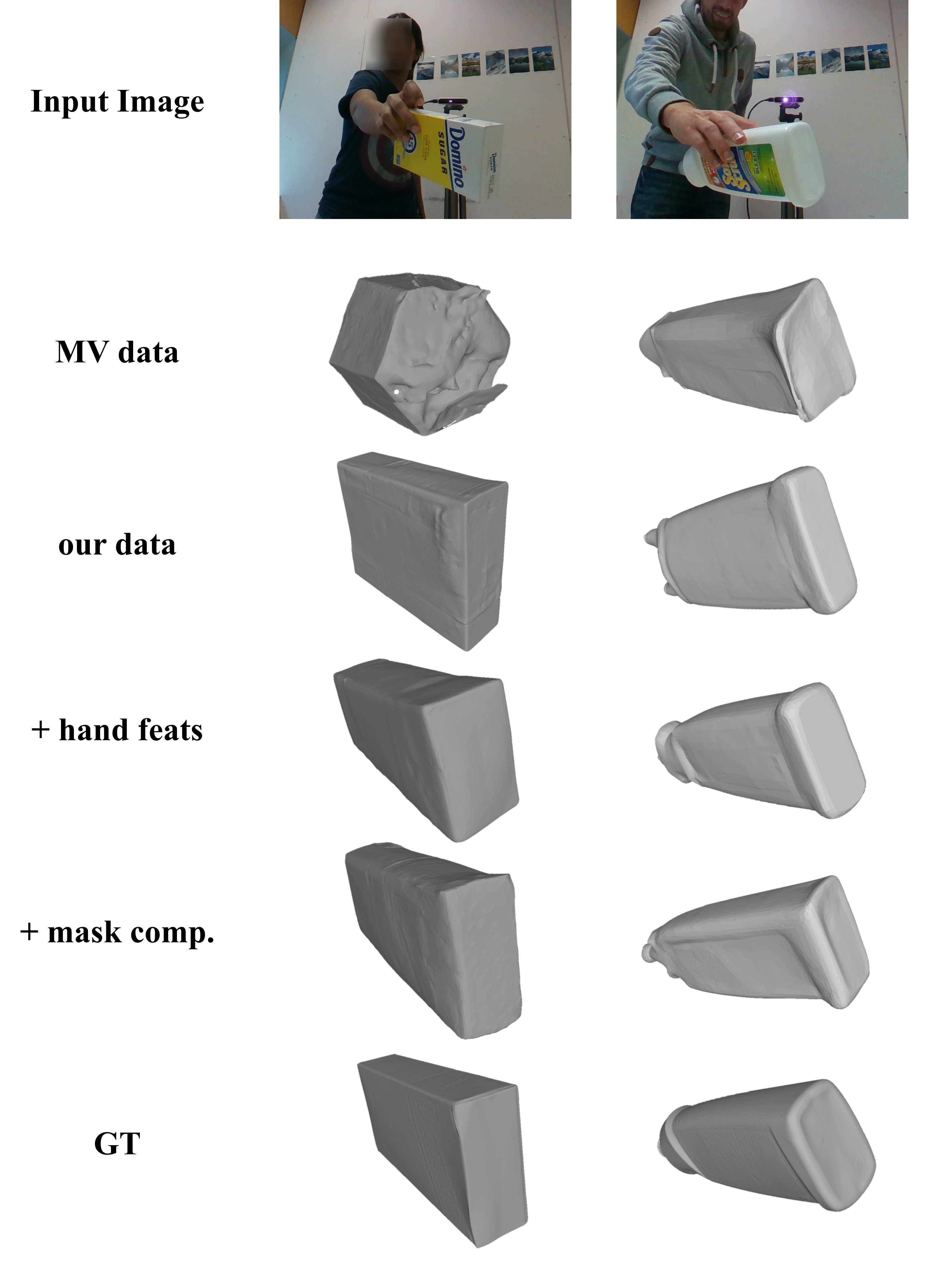}   
   \caption{\textbf{Qualitative comparisons for different variants of ForeHOI for ablative study on HO3D~\cite{hampali2020honnotate} dataset. Zoom
in for better visualization in detail.}}
   \label{fig:ablation}
   \vspace{-5mm}
\end{figure}

\subsection{Ablation studies}
\label{sec:ablation}
We ablate our two key designs: hand feature input and the 2D object mask completion branch of the pipeline, and the importance of our data in Tab.~\ref{tab:ablation}.

\noindent\textbf{The importance of our dataset} The problem of in-hand object reconstruction under dynamic occlusion can somewhat be viewed as a multi-view object reconstruction problem under occlusion: the camera movement with the object frozen is equal to the object movement under the object-camera coordinate. According to this assumption, we first trained our model on the commonly used object datasets~\cite{xiang2024structured}, which are easy to obtain in our early experiments. We add random masks to the multi-view images and select restricted views for each training step to imitate the hand-held environment. However, as shown in Tab.~\ref{tab:ablation}, the performance on the real-world dataset is very poor. Based on our observations, this is primarily attributable to two issues: Firstly, the semantic understanding and completion of 3D generative models rely on the object-centric coordinate system. When objects are rendered on a table, their orientations are generally aligned, whereas objects rotated by hands represent a completely unfamiliar scenario for models trained on normal data. Secondly, the use of random occlusions devoid of semantic information results in the failure to learn hand-object interactions. This insight has inspired two other key components in our framework design: the incorporation of hand inputs and mask completion, and also underscores the importance of our data.

\noindent\textbf{The influence of hand features} One of the key challenges in object reconstruction of HOI scenes is analyzing the hand-object relationship. The hand feature gives the model a direct understanding of where the occlusion comes from. Fig.~\ref{fig:ablation} shows that without the hand feature, the models sometimes fail to complete the object under hand contact areas with obvious fingerprint indentation.

\noindent\textbf{The influence of 2D mask completion branch} As previously mentioned, the masks caused by hand occlusion are relatively regular and contain information about hand-object interactions. After obtaining per-frame hand features, completing the full object mask becomes a relatively straightforward task. However, this simple task enhances the model's understanding of hand-object interactions, thereby significantly improving its performance in scenarios with substantial hand occlusion, as can be observed in Tab.~\ref{tab:ablation} and Fig.~\ref{fig:ablation}.

\begin{table}[htbp]
  \centering
    \begin{tabular}{c|ccc}
    \toprule
    Strategy & CD(cm) $\downarrow$ & \textcolor[rgb]{ .02,  .388,  .757}{F@5(\%)$\uparrow$} & \textcolor[rgb]{ .02,  .388,  .757}{F@10(\%)$\uparrow$} \\
    \midrule
    MV data  & 1.23  & 25.54 & 68.54 \\
    our data  & 0.91  & 53.64 & 80.63 \\
    + hand feats & 0.88  & 54.53 & 85.43 \\
    + mask comp. & 0.79  & 68.95 & 93.72 \\
    \bottomrule
    \end{tabular}%
    \caption{\textbf{MV data} means ReconViaGen~\cite{chang2025reconviagenaccuratemultiview3d} backbone trained on their dataset with random image masking. \textbf{our data} means ReconViaGen~\cite{chang2025reconviagenaccuratemultiview3d} backbone trained on our hand-object interaction data. Both \textbf{+hand feats} and \textbf{+mask comp.} represent versions trained on our data, with different architecture design.}
  \label{tab:ablation}%
  \vspace{-5mm}
\end{table}%

%% file: sec/5_conclusion.tex
\section{Conclusion}
\label{sec:conclusion}
\indent In this paper, we present ForeHOI, a novel feed-forward model for fast and accurate 3D object reconstruction from monocular HOI videos. Our work demonstrates that high-fidelity reconstruction of hand-held objects under severe occlusion can be learned end-to-end within a single network. This is achieved by jointly predicting 2D mask inpainting and 3D shape completion, allowing the two tasks to benefit from each other to generate more accurate final geometry. Our method eliminates the dependency on cumbersome pre-processing steps and maintains fast inference speed. Qualitative and quantitative results show that ForeHOI outperforms state-of-the-art methods on in-the-wild settings.
